\documentclass[letterpaper]{article}
\usepackage{aaai}
\usepackage{times}
\usepackage{helvet}
\usepackage{courier}
\usepackage{graphicx}
\usepackage{color}
\newcommand{\ml}[1]{#1}
\nocopyright

\title{RumourEval 2019: Determining Rumour Veracity and Support for Rumours}

\author{
Genevieve Gorrell\textsuperscript{1}, Kalina Bontcheva\textsuperscript{1}, Leon Derczynski\textsuperscript{2}, \\ \textbf{\Large{Elena Kochkina\textsuperscript{3}, Maria Liakata\textsuperscript{3}, and Arkaitz Zubiaga\textsuperscript{3}}}\\
\\
\textsuperscript{1}University of Sheffield, UK\\
\tt{(g.gorrell,k.bontcheva)@sheffield.ac.uk}\\
\\
\textsuperscript{2}IT University of Copenhagen, Denmark\\
\tt{ld@itu.dk}\\
\\
\textsuperscript{3}University of Warwick, UK\\
\tt{(e.kochkina,m.liakata,a.zubiaga)@warwick.ac.uk}
}

\begin{document}

\maketitle

\begin{abstract}
\begin{quote}
This is the proposal for RumourEval-2019, which will run in early 2019 as part of that year's SemEval event. Since the first RumourEval shared task in 2017, interest in automated claim validation has greatly increased, as the dangers of ``fake news'' have become a mainstream concern. Yet automated support for rumour checking remains in its infancy. For this reason, it is important that a shared task in this area continues to provide a focus for effort, which is likely to increase. We therefore propose a continuation in which the veracity of further rumours is determined, and as previously, supportive of this goal, tweets discussing them are classified according to the stance they take regarding the rumour. Scope is extended compared with the first RumourEval, in that the dataset is substantially expanded to include Reddit as well as Twitter data, and additional languages are \ml{also} included.
\end{quote}
\end{abstract}

\section{Overview}

\begin{table*}
\begin{center}
\begin{tabular}{|l|}
\hline
\\
\textbf{Veracity prediction. Example 1:}\\
\\
\textbf{u1:} Hostage-taker in supermarket siege killed, reports say. \#ParisAttacks –LINK– [true]\\
\\
\textbf{Veracity prediction. Example 2:}\\
\\
\textbf{u1:} OMG. \#Prince rumoured to be performing in Toronto today. Exciting! [false]\\
\\
\hline
\end{tabular}
\end{center}
\caption{Examples of source tweets with veracity value}
\label{tab:ex}
\end{table*}

Since the first RumourEval shared task in 2017 \cite{derczynski2017semeval}, interest in automated verification of rumours has only deepened, as research has demonstrated the potential impact of false claims on important political outcomes~\cite{allcott2017social}. Living in a ``post-truth world'', in which perceived truth can matter more than actual truth~\cite{dale2017nlp}, the dangers of unchecked market forces and cheap platforms, alongside often poor discernment on the part of the reader, are evident. For example, the need to educate young people about critical reading is increasingly recognised.\footnote{http://www.bbc.co.uk/mediacentre/latestnews/2017/fake-news} The European Commission's High Level Expert Group on Fake News cite provision of tools to empower users and journalists to tackle disinformation as one of the five pillars of their recommended approach.\footnote{http://ec.europa.eu/newsroom/dae/document.cfm?doc\_id=50271} Simultaneously, research in stance prediction and assembling systems to understand and assess rumours expressed in written text have made some progress over baselines, but a broader understanding \ml{of the relation between stance and veracity} -- and \ml{a more extensive} dataset -- \ml{are} required.

In a world where click-bait headlines mean advertising revenue, incentivising stories that are more attractive than they are informative, we are experiencing a deluge of fake news. Automated approaches offer the potential to keep up with the increasing number of rumours in circulation. Initial work~\cite{qazvinian2011rumor} has been succeeded by more advanced systems and annotation schemas~\cite{kumar2014detecting,zhang2015automatic,shao2016hoaxy,zubiaga2016analysing}. Full fact checking is a complex task that may challenge the resourcefulness of even a human expert. Statistical claims, such as "we send the EU £350 million a week", may offer a more achievable starting point in full fact checking, and have inspired engagement from researchers~\cite{vlachos2015identification} and a new shared task. FEVER;\footnote{https://sheffieldnlp.github.io/fever/} whilst this work has a different emphasis from rumour verification, it shows the extent of interest in this area of research. Other research has focused on stylistic tells of untrustworthiness in the source itself~\cite{conroy2015automatic,singhania20173han}. Stance detection is the task of classifying a text according to the position it takes with regards to a statement. Research supports the value of this subtask in moving toward veracity detection~\cite{ferreira2016emergent,enayet2017niletmrg}.

UK fact-checking charity Full Fact provides a roadmap\footnote{https://fullfact.org/media/uploads/full\_fact-the\_state\_of\_automated\_factchecking\_aug\_2016.pdf} for development of automated fact checking. They cite open and shared evaluation as one of their five principles for international collaboration, demonstrating the continuing relevance of shared tasks in this area. Shared datasets are a crucial part of the joint endeavour. Twitter continues to be a highly relevant platform, being popular with politicians. By including Reddit data in the 2019 RumourEval we also provide diversity \ml{in the types of users, more focussed discussions} and longer texts.

\subsection{Summary of RumourEval 2017}

RumourEval 2017 comprised two subtasks:

\begin{itemize}

\item{In subtask A, given a source claim, tweets \ml{in a conversation thread discussing the claim} are classified into support, deny, query and comment categories.}

\item{In subtask B, the source tweet that spawned the discussion is classified as true, false or unverified.}

\begin{itemize}

\item{In the open variant, this was done on the basis of the source tweet itself, the discussion and additional background information.}

\item{In the closed variant, only the source tweet and the \ml{ensuing} discussion were used.}

\end{itemize}

\end{itemize}

Eight teams entered subtask A, achieving accuracies ranging from 0.635 to 0.784. In the open variant of subtask B, only one team participated, gaining an accuracy of 0.393 and demonstrating that the addition of a feature for the presence of the rumour in the supplied additional materials does improve their score. Five teams entered the closed variant of task B, scoring between 0.286 and 0.536. Only one of these made use of the discussion material, specifically the percentage of responses querying, denying and supporting the rumour, and that team scored joint highest on accuracy and achieved the lowest RMSE. A variety of machine learning algorithms were employed. Among traditional approaches, a gradient boosting classifier achieved the second best score in task A, and a support vector machine achieved a fair score in task A and first place in task B. However, deep learning approaches also fared well; an LSTM \ml{-based approach} took first place in task A and \ml{an approach using} CNN took second place in task B, though performing less well in task A. Other teams used different kinds of ensembles and cascades of traditional and deep learning supervised approaches. In summary, the task attracted significant interest and a variety of approaches. However, for 2019 it is worth considering how participants might be encouraged to be more innovative in the information they make use of, particularly exploiting the output of task A in their task B approaches.

\subsection{How RumourEval 2019 will be different}

For RumourEval 2019 we plan to extend the competition through the addition of new data, including Reddit data, and through extending the dataset to include new languages, namely Russian~\cite{lozhnikov2018stance} and Danish.

In order to encourage more information-rich approaches, we will combine variants of subtask B into a single task, in which participants may use the additional materials (selected to provide a range of options whilst being temporally appropriate to the rumours in order to mimic the conditions of a real world rumour checking scenario) whilst not being obliged to do so. In this way, we prioritise stimulation of innovation in pragmatic approaches to automated rumour verification, shifting the focus toward success at the task rather than comparing machine learning approaches. At the same time, closed world entries are not excluded, and the task still provides a forum via which such approaches might be compared among themselves.

\subsection{Subtask A - SDQC support classification}

\begin{table*}
\begin{center}
\begin{tabular}{|l|}
\hline
\\
\textbf{SDQC support classification. Example 1:}\\
\\
\textbf{u1:} We understand that there are two gunmen and up to a dozen hostages inside the cafe under siege at\\
Sydney.. ISIS flags remain on display \#7News \textbf{[support]}\\
\hspace*{10mm} \textbf{u2:} @u1 not ISIS flags \textbf{[deny]}\\
\hspace*{10mm} \textbf{u3:} @u1 sorry - how do you know it’s an ISIS flag? Can you actually confirm that? \textbf{[query]}\\
\hspace*{20mm} \textbf{u4:} @u3 no she can’t cos it’s actually not \textbf{[deny]}\\
\hspace*{10mm} \textbf{u5:} @u1 More on situation at Martin Place in Sydney, AU –LINK– \textbf{[comment]}\\
\hspace*{10mm} \textbf{u6:} @u1 Have you actually confirmed its an ISIS flag or are you talking shit \textbf{[query]}\\
\\
\textbf{SDQC support classification. Example 2:}\\
\\
\textbf{u1:} These are not timid colours; soldiers back guarding Tomb of Unknown Soldier after today's shoot-\\
ing \#StandforCanada –PICTURE– \textbf{[support]}\\
\hspace*{10mm} \textbf{u2:} @u1 Apparently a hoax. Best to take Tweet down. [deny]\\
\hspace*{10mm} \textbf{u3:} @u1 This photo was taken this morning, before the shooting. [deny]\\
\hspace*{10mm} \textbf{u4:} @u1 I don’t believe there are soldiers guarding this area right now. [deny]\\
\hspace*{20mm} \textbf{u5:} @u4 wondered as well. I’ve reached out to someone who would know just to confirm\\
\hspace*{20mm} that. Hopefully get response soon. [comment]\\
\hspace*{30mm} \textbf{u4:} @u5 ok, thanks. [comment]\\
\\
\hline
\end{tabular}
\end{center}
\caption{Examples of tree-structured threads discussing the veracity of a rumour, where the label associated with each tweet is the target of the SDQC support classification task.}
\label{tab:exsdqc}
\end{table*}

Rumour checking is challenging, and the number of datapoints is relatively low, making it hard to train a system and to demonstrate success convincingly. Therefore, as a first step toward this, in task A participants track how replies to an initiating post orientate themselves to the accuracy of the rumour presented in it. Success on this task supports success on task B by providing information/features; for example, where the discussion ends in a number of agreements, it could be inferred that human respondents have verified the rumour. In this way, task A provides an intermediate challenge on which a greater number of participants may be able to gain traction, and in which a much larger number of datapoints can be provided. Table~\ref{tab:exsdqc} gives examples of the material.

\subsection{Subtask B - Veracity prediction}

As previously, the goal of subtask B is to predict the veracity of a given rumour, presented in the form of a post reporting an update associated with a newsworthy event. Given such a claim, plus additional data such as stance data classified in task A and any other information teams choose to use from the selection provided, systems should return a label describing the anticipated veracity of the rumour. Examples are given in table~\ref{tab:ex}. In addition to returning a classification of true or false, a confidence score should also be returned, allowing for a finer grained evaluation. A confidence score of 0 should be returned if the rumour is unverified.

\subsection{Impact}

RumourEval 2019 will aid progress on stance detection and rumour extraction, both still unbested NLP tasks. They are currently moderately well performed for English short texts (tweets), with data existing in a few other languages (notably as part of IberEval). We will broaden this with a multi-lingual task, having the largest dataset to date, and providing a new baseline system for stance analysis.

Rumour verification and automated fact checking is a complex challenge. Work in credibility assessment has been around since 2011~\cite{castillo2011information}, making use initially of local features. Vosoughi~\shortcite{vosoughi2015automatic} demonstrated the value of propagation information, i.e. the ensuing discussion, in verification. Crowd response and propagation continue to feature in successful approaches, for example Chen et al~\shortcite{chen2016behavior} and the most successful system in RumourEval 2017~\cite{enayet2017niletmrg}, which might be considered a contender for the state of the art~\cite{zubiaga2018detection}. It is clear that the two part task formulation proposed here has continued relevance.

Platforms are increasingly motivated to engage with the problem of damaging content that appears on them, as society moves toward a consensus regarding their level of responsibility. Independent fact checking efforts, such as Snopes~\footnote{https://www.snopes.com/}, Full Fact~\footnote{https://fullfact.org/}, Chequeado~\footnote{http://chequeado.com/} and many more, are also becoming valued resources. Zubiaga et al~\shortcite{zubiaga2018detection} present an extensive list of projects. Effort so far is often manual, and struggles to keep up with the large volumes of online material. It is therefore likely that the field will continue to grow for the foreseeable future.

Datasets are still relatively few, and likely to be in increasing demand. In addition to the data from RumourEval 2017, another dataset suitable for veracity classification is that released by Kwon et al~\shortcite{kwon2017rumor}, which includes 51 true rumours and 60 false rumours. Each rumour includes a stream of tweets associated with it. A Sina Weibo corpus is also available~\cite{wu2015false}, in which 5000 posts are classified for veracity, but associated posts are not available. Partially generated statistical claim checking data is now becoming available in the context of the FEVER shared task, mentioned above, but isn't suitable for this type of work. A further RumourEval would provide additional data for system development as well as encouraging researchers to compare systems on a shared data resource.

\begin{figure*}
\begin{center}
  \includegraphics[width=0.75\linewidth]{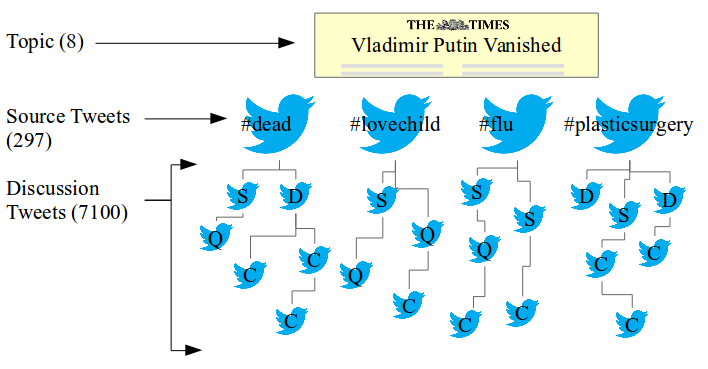}
  \caption{Structure of the first rumours corpus}
  \label{fig:corpus}
  \end{center}
\end{figure*}

\section{Data and Resources}

The data are structured as follows. Source texts assert a rumour, and may be true or false. These are joined by an ensuing discussion (tree-shaped) in which further users support, deny, comment or query (SDCQ) the source text. This is illustrated in figure~\ref{fig:corpus} with a Putin example.

The RumourEval 2017 corpus contains 297 source tweets grouped into eight overall topics, and a total of 7100 discussion tweets. This will become training data in 2019. We propose to augment this with at least the following:

\begin{itemize}

\item{New English Twitter test data}

\item{Reddit data in English}

\item{Twitter data in Russian}

\item{Twitter data in Danish}

\end{itemize}

Topics will be identified using Snopes and similar debunking projects. Potential source posts within these topics are identified on the basis of the amount of attention they attract; for tweets, number of retweets has been used successfully as an indicator of a good source tweet. Source texts are then manually selected from among these and labeled for veracity by an expert.

An existing methodology, used successfully in RumourEval 2017, allows us to harvest the ensuing discussions. The stances of discussion texts will be crowdsourced. Multiple annotators will be used, as well as testing, to ensure quality. Previous experience with annotating for this task shows that it can be achieved with a high interannotator agreement.

Twitter's developer terms, (Developer Policy, 1F.2a~\footnote{https://developer.twitter.com/en/developer-terms/agreement-and-policy}) state that up to 50,000 tweets may be shared in full via non-automated means such as download. This limit is sufficient for the tweets we envisage sharing. Twitter also requires that reasonable effort be made to ensure that tweets deleted by the author are also deleted by us (Developer Policy, 1C.3). To this end, the corpus will be checked for deleted tweets before release. In the event that Twitter requests a tweet be removed from the dataset, a new version of the data will be released to participants. It is unlikely that this would have a major impact on outcomes. Reddit places no restrictions on data redistribution.

\section{Evaluation}

In task A, stance classification, care must be taken to accommodate the skew towards the "comment" class, which dominates, as well as being the least helpful type of data in establishing rumour veracity. Therefore we aim to reward systems that perform well in classifying support, denial and query datapoints. To achieve this, we use macroaveraged F1, aggregated for each of these three types, and disregard the comment type entirely. Individual scores for the three main types will be provided separately in the final results.

In task B participants supply a true/false classification for each rumour, as well as a confidence score. Microaveraged accuracy will be used to evaluate the overall classification. For the confidence score, a root mean squared error (RMSE, a popular metric that differs only from the Brier score in being its square root) will be calculated relative to a reference confidence of 1. By providing these two scores, we give firstly a measure of system performance in the case of a real world scenario where the system must choose, and secondly a more fine-grained indicator of how well the system performed, that might be more relevant in the case that rumours are being automatically triaged for manual review. Note that it is possible for a system to score lower on accuracy but higher on RMSE compared with another system.

\section{Baseline}
For task A, we will provide code for a state-of-the-art baseline from RumourEval 2017 Task A ~\cite{kochkina2017turing} together with later higher-performing entry published at RANLP that year ~\cite{aker2017simple}. The latest state of the art system for stance classification on RumourEval 2017 Task A dataset ~\cite{veyseh2017temporal} may be provided in case of successful implementation.

For task B, we will provide our implementation of state-of-the-art baseline from RumourEval 2017 Task B ~\cite{enayet2017niletmrg} incorporating the best performing stance classification system.

\section{Task Organizers}

\textbf{Kalina Bontcheva}, University of Sheffield, UK. Research interests in social media analysis, information extraction, rumour analysis, semantic annotation, and NLP applications. Experience organising and running both workshops (including RDSM’2015) and conferences (the RANLP series, UMAP’2015). Email k.bontcheva@sheffield.ac.uk.

\textbf{Leon Derczynski}, IT University of Copenhagen, Denmark. Research interests in social media processing, semantic annotation, and information extraction. Experience running prior SemEval tasks (TempEval-3, Clinical TempEval, RumourEval) and major conferences (COLING). Email leod@itu.dk, telephone +45 51574948.

\textbf{Genevieve Gorrell}, University of Sheffield, UK. Research interests in social media analysis, text mining, text mining for health, natural language processing. Chaired the 2010 GATE summer school. Email g.gorrell@sheffield.ac.uk.

\textbf{Elena Kochkina}, University of Warwick, UK. Research interests in natural language processing, automated rumour classification, machine learning, deep learning for NLP. Experience participating in shared tasks, including RumourEval 2017. Email e.kochkina@warwick.ac.uk.

\textbf{Maria Liakata}, University of Warwick, UK. Research interests in text mining, natural language processing (NLP), biomedical text mining, sentiment analysis, NLP for social media, machine learning for NLP and biomedical applications, computational semantics, scientific discourse analysis. Co-chair of workshops in rumours in social media (RDSM) and discourse in linguistic annotations (LAW 2013). Co-organized RumourEval 2017. Email m.liakata@warwick.ac.uk.

\textbf{Arkaitz Zubiaga}, University of Warwick, UK. Research interests in social media mining, natural language processing, computational social science and human-computer interaction. Experience running 4 prior social media mining shared tasks including RumourEval 2017, and co-chair of workshops on social media mining at ICWSM and WWW. Email a.zubiaga@warwick.ac.uk.

\section*{Acknowledgements}

This work is supported by the European Commission’s Horizon 2020 research and innovation programme under grant agreement No. 654024, SoBigData.


\bibliographystyle{aaai}
\bibliography{proposal}

\begin{thebibliography}{}

\bibitem[\protect\citeauthoryear{Aker, Derczynski, and
  Bontcheva}{2017}]{aker2017simple}
Aker, A.; Derczynski, L.; and Bontcheva, K.
\newblock 2017.
\newblock Simple open stance classification for rumour analysis.
\newblock In {\em Proceedings of the International Conference Recent Advances
  in Natural Language Processing, RANLP 2017},  31--39.

\bibitem[\protect\citeauthoryear{Allcott and
  Gentzkow}{2017}]{allcott2017social}
Allcott, H., and Gentzkow, M.
\newblock 2017.
\newblock Social media and fake news in the 2016 election.
\newblock {\em Journal of Economic Perspectives} 31(2):211--36.

\bibitem[\protect\citeauthoryear{Castillo, Mendoza, and
  Poblete}{2011}]{castillo2011information}
Castillo, C.; Mendoza, M.; and Poblete, B.
\newblock 2011.
\newblock Information credibility on twitter.
\newblock In {\em Proceedings of the 20th international conference on World
  wide web},  675--684.
\newblock ACM.

\bibitem[\protect\citeauthoryear{Chen \bgroup et al\mbox.\egroup
  }{2016}]{chen2016behavior}
Chen, W.; Yeo, C.~K.; Lau, C.~T.; and Lee, B.~S.
\newblock 2016.
\newblock Behavior deviation: An anomaly detection view of rumor preemption.
\newblock In {\em Information Technology, Electronics and Mobile Communication
  Conference (IEMCON), 2016 IEEE 7th Annual},  1--7.
\newblock IEEE.

\bibitem[\protect\citeauthoryear{Conroy, Rubin, and
  Chen}{2015}]{conroy2015automatic}
Conroy, N.~J.; Rubin, V.~L.; and Chen, Y.
\newblock 2015.
\newblock Automatic deception detection: Methods for finding fake news.
\newblock {\em Proceedings of the Association for Information Science and
  Technology} 52(1):1--4.

\bibitem[\protect\citeauthoryear{Dale}{2017}]{dale2017nlp}
Dale, R.
\newblock 2017.
\newblock Nlp in a post-truth world.
\newblock {\em Natural Language Engineering} 23(2):319--324.

\bibitem[\protect\citeauthoryear{Derczynski \bgroup et al\mbox.\egroup
  }{2017}]{derczynski2017semeval}
Derczynski, L.; Bontcheva, K.; Liakata, M.; Procter, R.; Hoi, G. W.~S.; and
  Zubiaga, A.
\newblock 2017.
\newblock Semeval-2017 task 8: Rumoureval: Determining rumour veracity and
  support for rumours.
\newblock In {\em Proceedings of the 11th International Workshop on Semantic
  Evaluation (SemEval-2017)},  69--76.

\bibitem[\protect\citeauthoryear{Enayet and
  El-Beltagy}{2017}]{enayet2017niletmrg}
Enayet, O., and El-Beltagy, S.~R.
\newblock 2017.
\newblock Niletmrg at semeval-2017 task 8: Determining rumour and veracity
  support for rumours on twitter.
\newblock In {\em Proceedings of the 11th International Workshop on Semantic
  Evaluation (SemEval-2017)},  470--474.

\bibitem[\protect\citeauthoryear{Ferreira and
  Vlachos}{2016}]{ferreira2016emergent}
Ferreira, W., and Vlachos, A.
\newblock 2016.
\newblock Emergent: a novel data-set for stance classification.
\newblock In {\em Proceedings of the 2016 conference of the North American
  chapter of the association for computational linguistics: Human language
  technologies},  1163--1168.

\bibitem[\protect\citeauthoryear{Kochkina, Liakata, and
  Augenstein}{2017}]{kochkina2017turing}
Kochkina, E.; Liakata, M.; and Augenstein, I.
\newblock 2017.
\newblock Turing at semeval-2017 task 8: Sequential approach to rumour stance
  classification with branch-lstm.
\newblock {\em In Proceedings of SemEval.ACL}.

\bibitem[\protect\citeauthoryear{Kumar and
  Geethakumari}{2014}]{kumar2014detecting}
Kumar, K.~K., and Geethakumari, G.
\newblock 2014.
\newblock Detecting misinformation in online social networks using cognitive
  psychology.
\newblock {\em Human-centric Computing and Information Sciences} 4(1):14.

\bibitem[\protect\citeauthoryear{Kwon, Cha, and Jung}{2017}]{kwon2017rumor}
Kwon, S.; Cha, M.; and Jung, K.
\newblock 2017.
\newblock Rumor detection over varying time windows.
\newblock {\em PloS one} 12(1):e0168344.

\bibitem[\protect\citeauthoryear{Lozhnikov, Derczynski, and
  Mazzara}{2018}]{lozhnikov2018stance}
Lozhnikov, N.; Derczynski, L.; and Mazzara, M.
\newblock 2018.
\newblock {Stance Prediction for Russian: Data and Analysis}.
\newblock {\em arXiv preprint arXiv:1809.01574}.

\bibitem[\protect\citeauthoryear{Qazvinian \bgroup et al\mbox.\egroup
  }{2011}]{qazvinian2011rumor}
Qazvinian, V.; Rosengren, E.; Radev, D.~R.; and Mei, Q.
\newblock 2011.
\newblock Rumor has it: Identifying misinformation in microblogs.
\newblock In {\em Proceedings of the Conference on Empirical Methods in Natural
  Language Processing},  1589--1599.
\newblock Association for Computational Linguistics.

\bibitem[\protect\citeauthoryear{Shao \bgroup et al\mbox.\egroup
  }{2016}]{shao2016hoaxy}
Shao, C.; Ciampaglia, G.~L.; Flammini, A.; and Menczer, F.
\newblock 2016.
\newblock Hoaxy: A platform for tracking online misinformation.
\newblock In {\em Proceedings of the 25th international conference companion on
  world wide web},  745--750.
\newblock International World Wide Web Conferences Steering Committee.

\bibitem[\protect\citeauthoryear{Singhania, Fernandez, and
  Rao}{2017}]{singhania20173han}
Singhania, S.; Fernandez, N.; and Rao, S.
\newblock 2017.
\newblock 3han: A deep neural network for fake news detection.
\newblock In {\em International Conference on Neural Information Processing},
  572--581.
\newblock Springer.

\bibitem[\protect\citeauthoryear{Veyseh \bgroup et al\mbox.\egroup
  }{2017}]{veyseh2017temporal}
Veyseh, A. P.~B.; Ebrahimi, J.; Dou, D.; and Lowd, D.
\newblock 2017.
\newblock A temporal attentional model for rumor stance classification.
\newblock In {\em Proceedings of the 2017 ACM on Conference on Information and
  Knowledge Management},  2335--2338.
\newblock ACM.

\bibitem[\protect\citeauthoryear{Vlachos and
  Riedel}{2015}]{vlachos2015identification}
Vlachos, A., and Riedel, S.
\newblock 2015.
\newblock Identification and verification of simple claims about statistical
  properties.
\newblock In {\em Proceedings of the 2015 Conference on Empirical Methods in
  Natural Language Processing},  2596--2601.
\newblock Association for Computational Linguistics.

\bibitem[\protect\citeauthoryear{Vosoughi}{2015}]{vosoughi2015automatic}
Vosoughi, S.
\newblock 2015.
\newblock {\em Automatic detection and verification of rumors on Twitter}.
\newblock Ph.D. Dissertation, Massachusetts Institute of Technology.

\bibitem[\protect\citeauthoryear{Wu, Yang, and Zhu}{2015}]{wu2015false}
Wu, K.; Yang, S.; and Zhu, K.~Q.
\newblock 2015.
\newblock False rumors detection on sina weibo by propagation structures.
\newblock In {\em Data Engineering (ICDE), 2015 IEEE 31st International
  Conference on},  651--662.
\newblock IEEE.

\bibitem[\protect\citeauthoryear{Zhang \bgroup et al\mbox.\egroup
  }{2015}]{zhang2015automatic}
Zhang, Q.; Zhang, S.; Dong, J.; Xiong, J.; and Cheng, X.
\newblock 2015.
\newblock Automatic detection of rumor on social network.
\newblock In {\em Natural Language Processing and Chinese Computing}. Springer.
\newblock  113--122.

\bibitem[\protect\citeauthoryear{Zubiaga \bgroup et al\mbox.\egroup
  }{2016}]{zubiaga2016analysing}
Zubiaga, A.; Liakata, M.; Procter, R.; Hoi, G. W.~S.; and Tolmie, P.
\newblock 2016.
\newblock Analysing how people orient to and spread rumours in social media by
  looking at conversational threads.
\newblock {\em PloS one} 11(3):e0150989.

\bibitem[\protect\citeauthoryear{Zubiaga \bgroup et al\mbox.\egroup
  }{2018}]{zubiaga2018detection}
Zubiaga, A.; Aker, A.; Bontcheva, K.; Liakata, M.; and Procter, R.
\newblock 2018.
\newblock Detection and resolution of rumours in social media: A survey.
\newblock {\em ACM Computing Surveys (CSUR)} 51(2):32.

\end{thebibliography}

\end{document}